\documentclass{article}
\usepackage[table]{xcolor}
\usepackage{nips15submit_e,times}
\usepackage{url}
\usepackage{listings}
\usepackage{graphicx}

\title{Evolving imputation strategies for missing data in classification problems with TPOT}
\author{
  Unai Garciarena\thanks{\texttt{unai.garciarena@ehu.es}} \\
  Intelligent Systems Group  \\
Univ. of the Basque Country\\
 (UPV/EHU) \\
 San Sebastian, Spain \\
\And
  Roberto Santana\thanks{ \texttt{roberto.santana@ehu.es}}  \\
  Intelligent Systems Group \\
  Univ. of the Basque Country\\
 (UPV/EHU) \\
 San Sebastian, Spain \\
 \And
   Alexander Mendiburu \thanks{\texttt{alexander.mendiburu@ehu.es}}\\
   Intelligent Systems Group \\
  Univ. of the Basque Country\\
 (UPV/EHU) \\
 San Sebastian, Spain 
}

\nipsfinalcopy 

\begin{document}

\maketitle

\begin{abstract}
Missing data has a ubiquitous presence in real-life applications of machine learning techniques. Imputation methods are algorithms conceived for restoring missing values in the data, based on other entries in the database. The choice of the imputation method has an influence on the performance of the machine learning technique, e.g., it influences the accuracy of the classification algorithm applied to the data. Therefore, selecting and applying the right  imputation method is important and  usually requires a substantial amount of human intervention. In this paper we propose the use of genetic programming techniques to search for the right combination of imputation and classification algorithms. We build our work on the recently introduced Python-based TPOT library,  and incorporate a heterogeneous set of imputation algorithms as part of the machine learning pipeline search. We show that genetic programming can automatically find increasingly better pipelines that include the most effective combinations of imputation methods, feature preprocessing, and classifiers for a variety of classification problems with missing data. \\
  
{\bf{keywords}}: genetic programming, missing data, imputation methods, supervised classification
\end{abstract}

\section{Introduction}

Missing data can cause considerable losses in data quality, and therefore, information, in any kind of database (DB) in, potentially, any domain. This issue consists of a DB which does not contain all the information it is supposed to. There may be many reasons for this phenomenon, e.g., a human operator that failed to record a value, or a machine that was unable to transmit the values it recorded. Additionally, information that is not expected to be present can be identified as missing data if each case is not properly studied. Among others, this reason presents the study of missing data as a key to successfully prepare the data a given machine learning algorithm is fed with.
Imputation \cite{rubin_multiple_2004,little_statistical_2014} is one of the most used strategies for the missing data problem. These algorithms attempt to guess what the original, non-present, values were in the first place. To impute the missing data, the information available in the rest of the database (both in the defective observations and the complete ones) is used.

Previous studies have shown that the characteristics of missing data and those of the selected imputer are strongly related to information quality \cite{batista_analysis_2003,luengo_choice_2012}, as should be expected. However, no generic procedure providing a potential final user with the necessary knowledge to straightforwardly recognize a good imputer for a given dataset has been proposed.  Although the use of imputation methods is an essential requirement for the effective application of machine learning techniques, the choice of the right imputer is not trivial. The right choice of the imputation method will almost certainly depend on the type of classification problem at hand, or the distribution of the missing data.



In this paper we propose the use of a genetic programming (GP) \cite{koza_genetic_1990,langdon_foundations_2013,banzhaf_genetic_1998} strategy to search for the right combination of imputation and classification algorithms in such a way that the resulting pipeline can effectively deal with missing data. We build our work on the recently introduced Python-based TPOT library \cite{Olson_applications_2016}, which is able to evolve machine learning pipelines which consist of preprocessing, feature selection, and classification methods. While the preprocessing methods included in TPOT are diverse, imputation methods are not explicitly included.

We incorporate a heterogeneous set of imputation algorithms as part of the machine learning pipeline search. We inject missing data into a set of representative databases with a varying number of cases and variables, and show that the GP approach is able to consistently improve the classification accuracy along generations. Moreover, we investigate the relationship between the imputation methods and classification algorithms that appear in the best evolved pipelines. 

The article is structured as follows: In the next section we introduce the missing data problem and discuss imputation approaches.  Section~\ref{sec:CONTRIBUTION} presents our contribution, including a description of the TPOT library, the imputation methods we investigate and the algorithm used to simulate missing data in the complete databases.  Section~\ref{sec:EXPE} introduces the experimental framework used to evaluate our proposal and presents and discusses the numerical results. The main contributions of the paper are summarized in Section~\ref{sec:CONCLU}, which also includes some lines for future research.

\section{The missing data problem and the imputation approach} \label{def:IMPUT}

Missing data can be found in almost any kind of imaginable information collection. Different sources of missing data can result in different  \textit{missingness patterns} in the data. Most research works \cite{schafer_missing_2002,luengo_choice_2012} usually distinguish three divisions when it comes to classifying the arrangements described by missing values:

\begin{itemize}
	\item Missing Completely At Random (MCAR): When the probability of losing a value is unrelated to any information, i.e., when it is random, the missing data is cataloged as MCAR.
	\item Missing At Random (MAR): the MAR tag is used when a pattern can be identified, this is, we can find a common factor in all (or, at least a considerable amount of) the observations with missing values.
	\item Missing Not At Random (MNAR): This missing data type is similar to MAR. However, in this case the values \textit{causing} others to be missing are not known. This type is easily confused with MCAR.
\end{itemize}

The identification of the missing data pattern has proven to be key in data preprocessing. The distinctive characteristics in MCAR have made it a unique benchmark for imputation testing, since it has proven to show more contrast between imputation methods than other configurations \cite{garciarena_extensive_2017}. However, features in current databases tend to be interrelated. The result of this is that, even if missingness is caused by a reason usually linked to MCAR, the missing data class is relatively close to MAR.

Once we have correctly identified which one of the previous missing data types is the cause, one solution from the multiple ones available can be chosen. Depending on the characteristics of  the database, deleting incomplete observations could be an acceptable answer to the problem. An example of this is when there is very little defectiveness. However, in other cases, the data loss is not worth it. This is where imputation methods arise. 

\subsection{Imputation methods}

Since its formalization \cite{rubin_multiple_1978}, imputation has seen its status grow, and nowadays it is considered an essential preprocessing method  when operating with incomplete data \cite{batista_analysis_2003,luengo_choice_2012}. A common scenario for the application of imputation  is when the data is intended to undergo a machine learning process, whose capability to deal with data with absent entries is commonly limited.

Imputation methods are usually classified in two groups. The simple  and the complex ones. Simple methods correspond to straightforward operations, such as assigning the mean or median of each variable to all its lost items. When the percentage of lost entries is low, a simple method could be a good choice in terms of information loss avoidance and time consumption.

However, simple methods may not be a valid strategy when the amount of missing data for any of the variables rises. Computing the mean of a feature in such cases  would cause a significant reduction in the variance of the distribution of the data in the corresponding variable, which would result in a feature that contributes almost nothing when building a classification model. For this reason, more complex algorithms have been developed, with the main goal of producing more accurate results, taking into account the closeness to the original value. These more sophisticated methods are usually based on multiple imputation \cite{rubin_multiple_2004}, model creation \cite{honaker_amelia_2011},  or similarity computation \cite{andridge_review_2010}

\subsection{Imputation methods for classification}

Several works have studied the effect of imputation over a (mostly supervised) classification \cite{luengo_choice_2012,batista_analysis_2003}. These works usually select one or more databases to which missing data is introduced. Then the data is imputed, and used for classification. Finally, the imputation quality is measured based on the classification accuracy, computed via cross-validation. The results of these works usually suggest that complex imputation outperforms simpler strategies. For example, the study performed in \cite{luengo_choice_2012} investigated the relations between imputation methods and supervised classification algorithms. This work discerns three types of classifiers in rule inducting (e.g., decision trees), approximate models (SVM, MLP), and lazy (KNN) and applies a total of 23 classifiers on previously imputed databases treated by 14 imputation methods, from which 4 could be considered as simple strategies and the other 10 use a considerable amount of information and computational effort. The authors created a ranking based on the scores obtained from each database imputed with different methods. For the rule induction and approximate model classifiers (which include 19 out of 23 classifiers), any of the simple imputers failed to obtain a better rank than fourth, while for the lazy learning test obtained the best results. In \cite{luengo_study_2010}, a similar experiment was performed, but with a reduced set of classifiers (3). In this case, the experiment was performed over 22 databases, and, for the $3\times 22=66$ combinations, in only in 12 of them the simple methods obtained a better result, about $18\%$. 

An alternative approach to estimate quality of the imputer is to compare the imputed matrix to the original one using some distance measure \cite{troyanskaya_missing_2001,kim_missing_2005}. Generally, the normalized root mean square error is used. However, since the goal of this work is based on the classification algorithm performance, these approaches are out of scope.

\subsection{Genetic programming}

Genetic Programming (GP) was introduced in the early 90's \cite{koza_genetic_1990} as a paradigm of automatically ``creating" computing solutions for any class of problems. Instead of having a human-developed program, this concept proposes to seek for a solution in a search space where all possible programs can be found. This idea implements an evolutionary algorithm \cite{whitley_genetic_1994} to perform such exploration, considering evaluable programs as individuals (shaped as trees), and contemplating typical genetic operators such as crossover and mutation. GP has proven to be a good alternative to human knowledge when it comes to the automatic generation of programs, as it has sometimes produced programs as good as human-made versions, or has even improved classical program developments and implementation processes \cite{koza_annual_2017}.

\subsection{Related Work}

There have been previous approaches that apply GP to design imputation methods. For example, in \cite{tran_genetic_2016}, a regression imputation method is developed using GP. They point out the suitability of GP to design a symbolic regression procedure since it is a non-parametric regressive model, which seeks for both an optimal structure and parameters. The method was compared to other three techniques based on regression and two tree-based algorithms, and tested regarding the accuracy that different classifiers obtained after being trained with data treated by the imputation methods. The used classifiers were K Nearest Neighbors, Na\"ive Bayes, Support Vector Machine and Multi Layer Perceptron. The results showed that the proposed method consistently offered significantly better performance compared to the selected benchmark.

In \cite{tran_multiple_2015}, the same authors  proposed a similar method, with a more explicit multiple imputation implementation, and this time they used the accuracy and the error between the imputed and the real values as a metric to evaluate the imputation method. Again, the proposed method regularly beat the other methods.

\cite{de_resende_time_2016} focuses its efforts on time-series data. The authors exploit this characteristic by firstly imputing the dataset with Lagrange interpolation, and then perform a regression function computation with a genetic programming approach. The results of the method are compared to other IMs, such as mean, cluster, and KNN imputation, and only KNN seemed to obtain better results in some occasions. Anyhow, the proposed method offered the best results more frequently.

Notice that the approach we propose in this paper is different. We evolve combinations of different steps involved in the classification process. We improve these combinations by adding the possibility to contain imputation methods within them.

\section{Evolving automatic ML pipelines for databases with missing data} \label{sec:CONTRIBUTION}

In this paper we address  the question of how to design a machine learning solution, involving different modules, for classification problems with missing data. In these problems, a database of observations (or cases), each of which corresponds to a row in the database, is given. Each observation has an associated class. The general goal is to learn to predict the class solely using the information contained in the observations. Generally, classifiers require the observations to be complete. Therefore, a solution to the missing data problem would include the choice of an imputation method. Furthermore, realistic machine learning pipelines also add steps in which the data is preprocessed in other ways (e.g., PCA and variable normalization), or subsets of features are selected from the original data. 

\begin{figure}
	\begin{center} 
		\scalebox{0.3}{\includegraphics{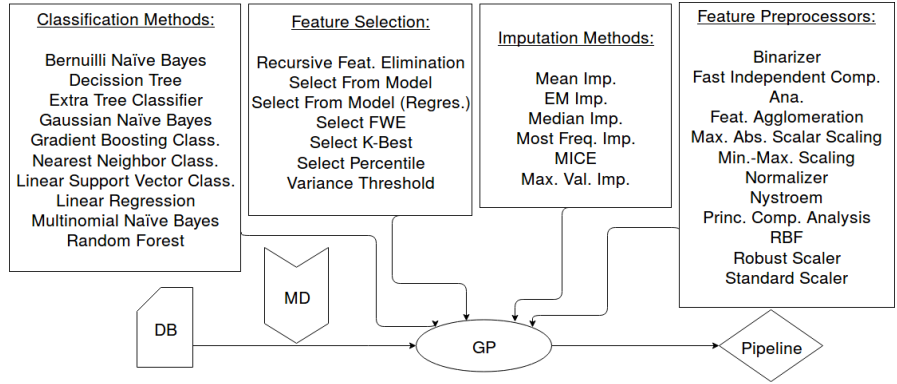}} 
		\caption{TPOT Workflow} 
		\label{fig:IMPUTATION} 
	\end{center} 
\end{figure}

There are several possible combinations of imputation, preprocessing, feature selection and classification methods. For a given database, trying all possible combinations to identify the one that maximizes a given criterion, e.g., the classification accuracy, is often unfeasible. Search-based methods are a feasible alternative in this scenario. Therefore, we focus on automatically generating a machine learning pipeline for maximizing the classification accuracy in problems with missing data. Figure~\ref{fig:IMPUTATION} shows a diagram of the different components involved in our approach to this problem. Three of these components are already included in the TPOT implementation (Classification method, feature selection, and preprocessors). Our contribution lies in the design and implementation of the fourth one (Imputation methods).


We organize the presentation of our proposal in three steps: First, we describe the TPOT library, the work in which our approach is inspired. Secondly, we explain the characteristics of the imputation methods we implemented, which are the building blocks of our approach. Thirdly, we introduce the method conceived to inject missing data into the original databases.

\subsection{TPOT, a platform for automatic parameter selection}

TPOT (Tree-based Pipeline Optimization) \cite{Olson_applications_2016} progressively evolves machine learning pipelines for regression and classification problems. The program implements a multi-objective GP algorithm \cite{barlow_incremental_2004,bleuler_multiobjective_2001} where the programs correspond to pipelines of the popular scikit-learn (sklearn) \cite{pedregosa_scikit-learn:_2011} library. 

TPOT defines all the machine learning techniques it uses as GP primitives, and organizes them in a tree structure. These constructions are the elements evolved by the underlying genetic component in the core of the project.

The other component of the TPOT project is the DEAP (Distributed Evolutionary Algorithms in Python) library \cite{fortin_deap:_2012}. This package endows TPOT with the method to implement genetic programming over the sklearn-based pipelines. DEAP is basically a framework that simplifies the task of creating fast evolutionary prototypes, which perfectly suits TPOT. The contribution of DEAP on this specific task of serving TPOT is explained in the following paragraphs.

The overall workflow is not far from the standard GP procedure. As an initial step, a population is randomly initialized, where each individual is a tree-shaped sklearn pipeline. Next, TPOT evaluates every tree regarding the accuracy that the components inside the tree managed to obtain, computed via balanced 3-fold cross-validation. Then, the best scoring individuals are selected (via the NSGA2 strategy \cite{deb_fast_2002}) to take part in the creation of the next population. For this step, the top trees are susceptible to being subjected to crossover and mutation. Once this sub-process is finished, the whole method is rerun. The algorithm stops when the number of specified generations is reached. Every generation, the algorithm updates a ``\textit{hall of fame}", a Pareto front 
comprising all solutions in every generation.

As previously stated, GP, as implemented in TPOT,  pursues two objectives rather than simply maximizing the classification accuracy. TPOT also attempts to keep trees as simple as possible. This constraint addresses the necessary issue of regulating the complexity of the final solution, resulting in a potentially considerable computation-time gain. Without this restriction, the algorithm could generate intractable pipelines, composed by heaps of sklearn methods. Obviously, this multi-objective feature generates a Pareto front as an output, from which TPOT selects the individual with the highest accuracy. Therefore, even though the second objective has had a relevant role during the TPOT general run, it remains unused in the last selection.

Despite being recently introduced software, TPOT has been successfully applied to different problems \cite{Olson_applications_2016,olson_tpot:_2016,elizalde_experiments_2016}. In \cite{Olson_applications_2016}, it was tested using artificial data sets and the (real-world) CGEMS Prostate Cancer Data Set.  In \cite{olson_tpot:_2016}, it was compared to a random forest classifier over $150$ real databases from different sources. TPOT obtained significantly better results than the random forest classifier in $21$ of them, and significantly worse in $4$. Finally, for $125$, no significant differences were found.

The following code shows an example product of TPOT, a sklearn representation for a program that includes an imputation method, a preprocessing step and a classifier with different parameters:

\begin{lstlisting}
   Pipeline(steps=[(`mfimputer', MFImputer()), 
      (`maxabsscaler', MaxAbsScaler(copy=True)),
      (`multinomialnb', MultinomialNB(alpha=1, fit_prior=True))])
\end{lstlisting}

And how the same expression would be coded internally in the program, before being processed by sklearn:

\begin{lstlisting}
   MultinomialNB(MaxAbsScaler(MFImputer(input_matrix), 
      MaxAbsScaler__copy=True), MultinomialNB__alpha=1, 
      MultinomialNB__fit_prior=True)
\end{lstlisting}

This is a \textit{strongly typed} string, since the parameters are individually and explicitly linked to specific parameters of specific functions. This is necessary, for example, to introduce parameters of the required types, and avoid entering a Boolean where an array is expected. However, this notation is not practical, and it is therefore converted to a more tractable notation to undergo processes such as genetic operations or conversion to sklearn pipeline: 

\begin{lstlisting}
   [`MultinomialNB', [`MaxAbsScaler', [`MFImputer', 
      `input_matrix'], True], 1, True]
\end{lstlisting}	

As it can be seen the trees are coded as lists of lists (brackets denote lists in python syntax). In this case, ``input\_matrix" would be our data, which would firstly be treated by the ``MFImputer". Then ``MaxAbsScaler" would be applied to the imputed matrix, with ``True" as the argument for the ``copy" parameter. Finally, the data would be used for classification, via ``MultinomialNB", with ``1", and ``True", as parameters for the alpha, and prior class probability learning (\textit{fit\_prior}), respectively.

DEAP straightforwardly applies simple operators, such as one-point-crossover, over this kind of lists. For instance, if the algorithm decided to mate this input with another similar one, it would simply select a node and cross it with another node with similar characteristics from another tree.

For this instance, we will be crossing the preprocessing nodes, with the imputer within it; ``MaxAbsScaler" (with ``MFImputer") and ``FastICA" (with ``MICEImputer"):

\begin{lstlisting}
   [`MultinomialNB', [`MaxAbsScaler', [`MFImputer', 
      `input_matrix'], True], 1, True]
   ['KNeighborsClassifier', ['FastICA', ['MICEImputer', 
      'input_matrix', 11, 6], 1.0], 45, 1, 'distance']
\end{lstlisting}

for resulting in:

\begin{lstlisting}
   [`MultinomialNB', ['FastICA', ['MICEImputer', 
      'input_matrix', 11, 6], 1.0],  1, True]
   ['KNeighborsClassifier', [`MaxAbsScaler', [`MFImputer', 
      `input_matrix', None], True], 45, 1, 'distance']
\end{lstlisting}

Note that this operation could be performed with any nodes, as long as the individuals keep matching with the grammatic defined in TPOT.

This kind of operation can also be performed with other altering strategies, i.e., mutation. In this case, we modify a terminal node of the tree (a parameter of a method) to another feasible value:

\begin{lstlisting}
   [`MultinomialNB', [`MaxAbsScaler', [`MFImputer', 
      `input_matrix'], True], 1, True]
\end{lstlisting}

In this example, we modify the node representing the ''alpha" parameter in ``MultinomialNB" (the first one) from 1 to 10, another selection in the range of plausible values:

\begin{lstlisting}
   [`MultinomialNB', [`MaxAbsScaler', [`MFImputer', 
      `input_matrix'], True], 10, True]
\end{lstlisting}

\subsection{Implementing imputation methods}

Instead of exhaustively evaluating combinations of imputation methods and classifiers over databases with missing values, we exploit TPOT to search for the most effective combination. We recall that there is an important difference between preprocessing methods as implemented in TPOT, and imputation methods.  The former are conceived for transforming complete data. Imputation methods are oriented to replace the missing data and, when used in the wrong way, they can introduce important biases in the information. In these cases, the results of the classification process are not only influenced by the quality of the classifier but also by the adequacy of the imputation method. This highlights the relevance, and also the difficulty, of incorporating the imputation analysis to the automation of the ML pipeline. 

In our implementation, we evaluate six imputation methods. Firstly, four simple algorithms; mean, median and most frequent which are self explaining, and are already implemented in sklearn, and max. imputer, which computes the maximum value of a variable from the available data and imputes it on all the missing parts. In addition, we incorporate two complex imputation methods that have been extensively applied as a previous step for classification.


MICE \cite{buuren_mice:_2011} (Multiple Imputation based on Chained Equations) clearly differentiates two ideas on its operations. Firstly, it performs a simple imputation (e.g., mean) with the objective of re-imputing each missing value with a regression method until a certain threshold is met. In other words, variables are imputed with a straightforward method, to later use the complete DB to re-impute each originally missing values via regression. Multiple Imputation is the other concept influencing the design of the method, as the previously explained re-imputation procedure is performed $m$ times, resulting in so many outcomes. A final product is computed from all the primary imputations. This method has proven to be top class in missing data restoration when that information is going to be fed to a machine learning algorithm, regarding classification accuracy \cite{ambler_comparison_2007}. However, the computational cost could be considered an enormous drawback when time is a key aspect.

The EM (Expectation-Maximization) method \cite{honaker_amelia_2011} is also the consequence of the combination between a base method and multiple imputation. To start with, an EM approach is used to compute various predefined statistical parameters from the known data, which is assumed to have a normal distribution. Similarly to the previous method, the originally missing values are re-imputed with models built from mentioned parameters, iteratively.  Once again, this algorithm is operated multiple times, and the results are combined similarly. This method also produces relatively good imputation, even if it is usually outperformed by MICE. Nevertheless, the elapsed time is considerably smaller compared to MICE, which results in a much better imputation quality/elapsed time ratio.

These two complex methods have been implemented using two \textbf{R} packages, each one respectively described in the two aforementioned references to the algorithms.

\subsection{Injecting missing data in the database}

Previous studies \cite{garciarena_extensive_2017} have shown that MCAR missing data type introduction maximizes the contrast produced by posterior imputation and supervised classification. For this reason, this missing data configuration is the most suitable one to measure the imputation performance in this case. For this instance, we randomly selected certain positions in the complete database, and changed their values to ``NaN". Halting was conditioned to a certain percentage of missingness, over the total quantity of values in the database.

\section{Experiments} \label{sec:EXPE}

In this section, we validate the results of our approach. The goals of our experiments are twofold. First, we would like to investigate whether the evolved pipelines are able to produce high classification accuracies even when missing data is introduced. To this end, we compare the accuracies obtained by our approach to those produced by TPOT over the complete databases. Our expectation is that missing data will produce a deleterious effect in the accuracy. Still, we expect this effect to be ameliorated by the use of imputation methods. We will assume that deleting observations with missing data is not a general and realistic approach, since, even if a dataset has a small percentage of MD and the incomplete cases are well distributed among the cases, removing observations can produce significant decrease in the number of available overall information.

The second goal of our experiments is trying to shed some light over the problem of discovering relations between imputation and classification. We analyze the best solutions produced by the introduced approach to determine  whether there exist dependencies between imputation and classification methods at the time of producing highly accurate pipelines. This question is relevant in order to produce general recommendations at the time of combining imputers and classifiers. 

This section is organized as follows. The next subsection presents a global view of the experimental design. Subsequently, the benchmark of classification problems and the parameters of the GP approach are presented. The two closing subsections respectively address the above-mentioned research questions, presenting the numerical results of the algorithms and discussing them.

\subsection{Experimental design} \label{sec:design}

The general work-flow of the experiment is described in the following steps:

\begin{enumerate}
	\item Missing data is introduced into the selected database.
	\item The incomplete DB is split into two databases (DB-Train and DB-Test). We use a random and unique partition of the dataset through all experimentation.
	\item DB-Train is used to evolve a pipeline using TPOT.
	\item The best pipeline among those evolved by TPOT is tested using DB-Test. This is performed in order to obtain an accuracy, which is regarded as the score of the pipeline. We use accuracy to remain consistent with both, the TPOT metric system, and the metric used in \cite{olson_evaluation_2016}.
	\item The best pipeline is stored along with its corresponding score, in addition to the other pipelines and accuracies generated from the rest of the benchmark.
\end{enumerate}

We notice that using a unique partition of the database could eventually bias the results of the comparison to the specific characteristics of the training set. Varying the train/test partition could be used to increase the accuracy of our experimentation but a larger set of experiments would be required to account for this extra source of variability, and this was not possible due to computational constraints.

The complete experimental process was repeated $20$ times to avoid biased results, which could be produced due to the stochastic component of the GP algorithm.

TPOT provides a natural framework to address the pipeline optimization problem. Using other optimization algorithms would require the design of specialized operators to search in the same space of pipelines. Additionally, TPOT uses an internal tree representation for its individuals, which suits perfectly genetic programming, resulting in a more than appropriate framework to make our contribution on.

\subsection{Benchmark classification problems and parameters of the GP approach}

This design was applied to a set of $23$ databases of those available from  \url{http://www.randalolson.com/data/benchmarks/}. These databases were originally used in \cite{olson_benchmark_2017} to investigate the performance of TPOT. The original  collection contains $151$ databases, ranging from 420 Bytes to  628 Mega Bytes. From this original collection we arbitrarily chose $23$ (all the databases starting with the letter ``a"), which range from 680B to 2.5MB, resulting in a set that contains float and integer variables, and combines binary and multi-class databases.

In our experiments, the GP algorithm used a population of $100$ individuals,  and a maximum of $50$ generations. The choice of the number of evaluations was due to the computational time constraints in the experiments. The percentage of missing data introduced was $7\%$ as in \cite{garciarena_extensive_2017}. This way, we can also roughly compare the effectiveness of the search algorithm compared to a full search. The initial database split (see Section \ref{sec:design}) is set at $75\%$. Moreover, this experimentation did not consider the XGBoost classifier, which is optional for TPOT. 

Additionally, we performed the same experimentation with the same parameters once more, only this time missing data was not introduced, and, obviously, the imputation operators were not used to evolve the pipelines. In other words, the experiments were also run with regular TPOT.

\subsection{Influence of the missing data on the accuracy results}

The goal of the first study  is to compare the pipelines generated using the original TPOT (without the addition of imputation methods) and our proposal, for two different types of databases: with and without missing values. We expect that injecting missing data will degrade the performance of the classifiers. However, we would like to measure how this degradation is alleviated by the use of imputation methods. 

\begin{figure}
	\begin{center} 
		\scalebox{0.32}{\includegraphics[trim={0 7cm 4cm 0}, clip]{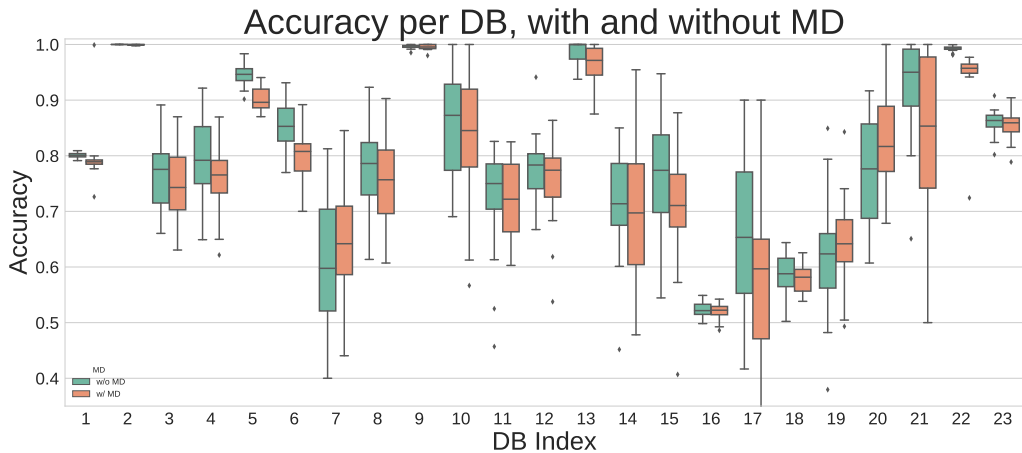}} 
		\caption{Box plot produced using the accuracies obtained using the top pipelines in each one of the $20$ runs of the experiment, for each DB. This plot also discerns databases that had no missingness injected and corresponding databases with missing values introduced and imputed.} 
		\label{fig:Acc} 
	\end{center} 
\end{figure}

The outcome of the comparison is summarized in Figure~\ref{fig:Acc}. This box-plot compares, for each database, the accuracies produced by the best evolved pipelines, discerning the results obtained with and without missing data injection. Note that, for this, and the subsequent figures, the database names have been substituted by indexes, in alphabetical order.

As expected, pipelines evolved from databases with missing values performed worse, overall. However, there are some exceptions. For example, pipelines developed from DB7 and DB20 obtained \textit{better} results generally across all the $20$ runs of the experiment.

Finally, some configurations involving missing values have managed to produce surprisingly good results, as proven by the top outliers in databases $1$ and $19$.

To determine if differences described in Figure \ref{fig:Acc} were statistically significant, the 20 outcomes run over each incomplete database were subjected to a statistical test, in which they were compared to their counterpart obtained from that same database, in its complete version. Table \ref{tab:dunn} summarizes the results of the tests.

\begin{table}[h]
	\scriptsize
	\begin{center}
		\begin{tabular}{|c|c|c|c|c|c|c|c|c|c|c|c|} 	\hline
			DB 1& DB 2& DB 3& DB 4& DB 5& DB 6& DB 7& DB 8& DB 9& DB 10& DB 11& DB 12\\ 
			
			\cellcolor{cyan!25}0.0& \cellcolor{cyan!25}0.0& 0.26& \cellcolor{cyan!25}0.05& \cellcolor{cyan!25}0.0& \cellcolor{cyan!25}0.0& 0.15& 0.15& 0.29& 0.37& 0.27& 0.33\\
			\hline
		\end{tabular}
		\begin{tabular}{|c|c|c|c|c|c|c|c|c|c|c|}
			\hline
			DB 13& DB 14& DB 15& DB 16& DB 17& DB 18& DB 19& DB 20& DB 21& DB 22& DB 23\\
			\cellcolor{cyan!25}0.01& 0.3& \cellcolor{cyan!25}0.08& 0.31& \cellcolor{cyan!25}0.05& 0.18& 0.16& 0.1& \cellcolor{cyan!25}0.02& \cellcolor{cyan!25}0.0& 0.19 \\
			\hline
		\end{tabular}
		\vspace{.5cm}
		\caption{p-values generated by applying the Dunn statistical test to each pair of database accuracies, over all 20 runs. Samples that generate a p-value under 0.05 are considered to be significantly different. A blue cell represents a database that generated remarkably better results on its original version compared to its incomplete variant. A non-highlighted cell means that there were found no differences between accuracies generated from the two versions of the database.}
		\label{tab:dunn}
	\end{center}
\end{table}  

\subsection{On the right combination of imputation methods and classification algorithms}

Once the effect of missing data has been analyzed, the next step focuses on determining whether our approach can be useful to identify how imputers and classifiers interact. To this end, we analyze the results on incomplete databases.

\begin{figure}
	\centering
	\scalebox{0.5}{\includegraphics[trim={0 0 31cm 0}, clip]{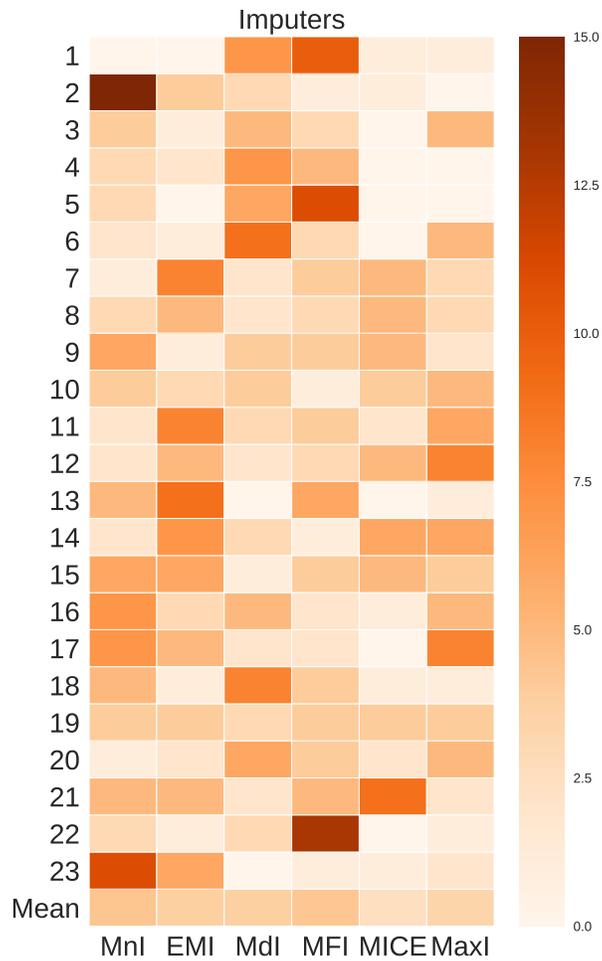}} 
	\caption{Frequency of imputers in the best evolved pipelines for each database. As it can be seen in the color scale, the darker the color, the more appearances a method had in an optimal pipeline.} 
	\label{fig:Imp} 
	
\end{figure}
\begin{figure}
	\centering
	\scalebox{0.5}{\includegraphics[trim={0 5cm 29cm 0}, clip]{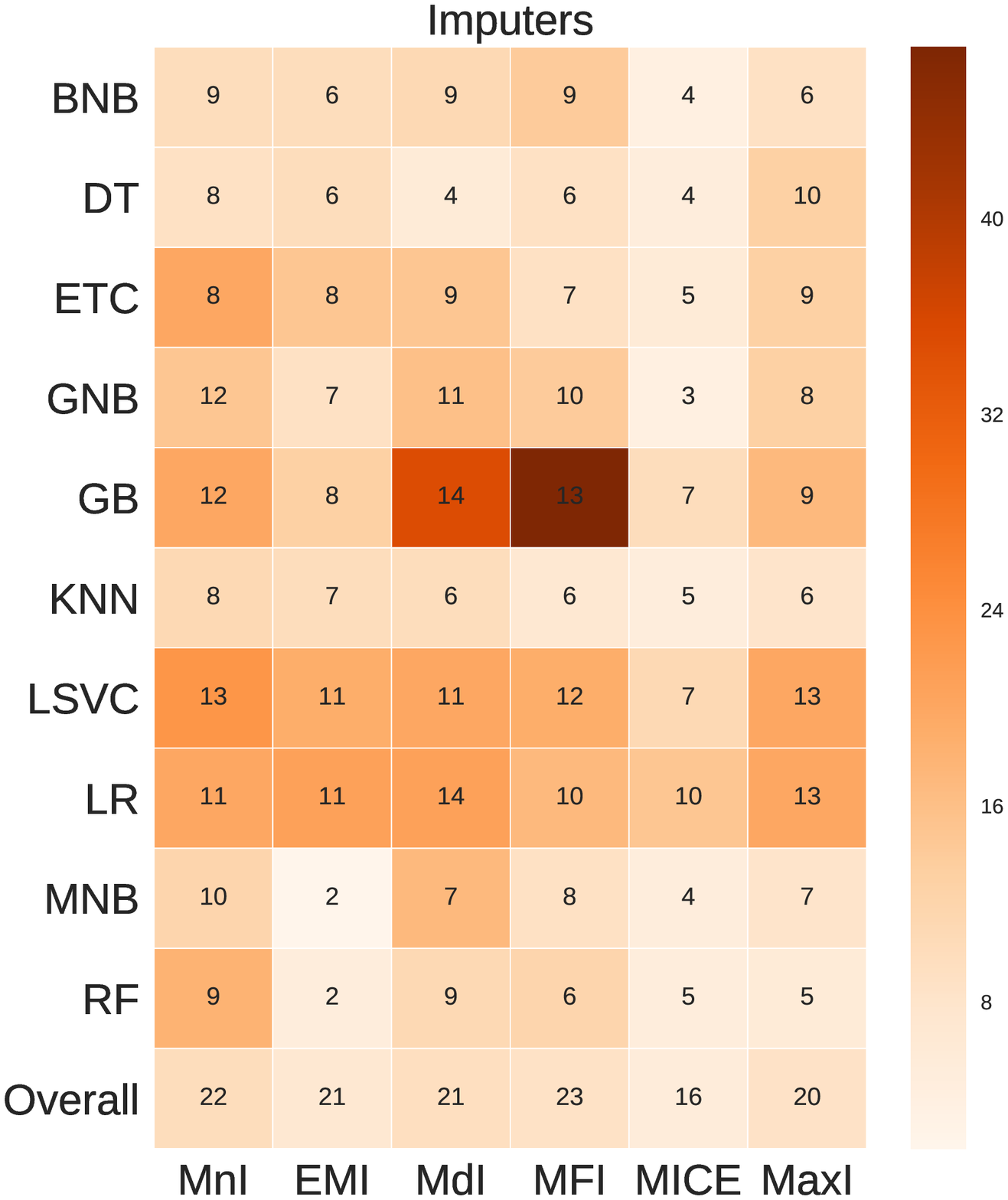}} 
	\caption{Average appearance rate for imputer/classifier pairs in the best evolved pipelines. Darker color correspond to higher appearance frequency. Number refers to the amount of different databases (out of 23) where each pair appears in at least one of the pipelines.} 
	\label{fig:Rel} 
\end{figure}

Figures \ref{fig:Imp} and \ref{fig:Rel} show heatmaps where darker colors correspond to higher appearance frequency. Note that the figures show compressed names for imputers and classifiers. Full names can be found in Figure \ref{fig:IMPUTATION}. Figure \ref{fig:Imp} represents the amount of times an imputer was part of the best pipeline generated for a specific database in all 20 runs of the experiment. Figure \ref{fig:Rel} represents the amount of times an imputer/classifier pair was comprised  in the best pipelines. Figure \ref{fig:Rel} also displays a number, which stands for the amount of different databases it was found in, i.e., a combination would only get one point for each database, even if it was present in more than one run for the same data. These numbers serve as a descriptor of how robust a particular combination of imputers and classifiers was at the time of dealing with missing data in a diverse set of databases.

An analysis of Figure \ref{fig:Imp} reveals the existing uncertainty when choosing imputation methods. Even though $23$ databases were considered, only two of them showed preference for one IM, appearing the latter in more than $11$ occasions. Another remarkable fact resides in the good performance offered by the simple imputation methods, as opposed to the usual result lines presented in the state-of-art.

In terms of the global behavior of the classifiers, Figure~\ref{fig:Rel} shows that the Gradient Boosting classifier produces good results regardless of the imputation method it is combined with, except for EM and Mean.

\subsection{Discussion}

In Figure \ref{fig:Acc}, we can see how overall databases with no missing data introduced offered better classification accuracies than their incomplete versions. However, if we take a closer look, some pipelines that obtained better performance generated from databases with missing values introduced and imputed stand out (DB7 is a clear example). At a first glance, this behavior could seem indicative of \textit{good} imputation, since the classification accuracy has increased. Nevertheless, in those cases imputation produced overconfidence on the classifier, which makes the model believe that there is less uncertainty than there really is, potentially leading to miss-classification. This is why several runs of the GP algorithm are necessary, not to be misled by uncommon behaviors and outliers, such as the one that obtained a 100$\%$ accuracy in database 1.

In any case, imputation seems to have softened the effects of missing values. Overall, databases that generated non-top accuracies had similar results between original and incomplete versions of data, and only some of the top accuracies (about $90\%$) offered significantly different results (DB5 and DB22). The fact that in most cases the generated accuracies from both complete and incomplete databases vary in similar ranges makes us believe that imputation is a more-than-valid strategy to consider when facing a missing data problem.

Results in these figures are statistically supported by Table \ref{tab:dunn}. This table confirms that 10 out of 23 databases showed significantly disparate results. The fact that the other 13 did not show these differences backs up the theory that, even if missing values harm data quality, imputation methods can be considered a viable solution. 

Concerning the information represented in Figure \ref{fig:Imp}, there is not clear indication that an imputation method clearly and consistently outperforms the other ones. More than one strategy had a major scoring difference over the rest in at least one database, (MICE on 21, most frequent on 22, and mean on 2, for example). This backs up what other works of the state-of-the-art conclude, that there is not a universally better imputation method regarding posterior classification accuracy.

Nevertheless, a quick analysis of the previous figure indicates most frequent imputation as the method that at the end of the process generated the best accuracies, overall.

This observation is partially supported by Figure \ref{fig:Rel}. In this case, even though most frequent imputation is repeatedly found in the best evolved pipelines, it is median imputation which, combined with different classification methods, appears in more distinct databases. However, all imputation methods were present in most part of database pipelines, at least once. We can see that Gradient Boosting benefited median and most frequent methods when combined, but in other cases, it is median imputation which clearly outperforms the most frequent strategy.

One possible explanation of these results is that they could be product of the small amount of missing data introduced, since the performance of the simple methods in some cases tends to drastically decrease as the missingness increases, while more sophisticated methods suffer these effects in a softer manner \cite{acuna_treatment_2004}.

\section{Conclusions} \label{sec:CONCLU}

Automatic generation of machine learning pipelines is one of the relevant topics to extend the scope of application and efficiency of machine learning software. Missing data is a frequent difficulty found at the time of applying machine learning algorithms. In this paper we have proposed a framework to automatically select imputation methods as part of classification pipelines.

We have also investigated the relationship between imputers and classifiers using an approach, that to the knowledge of the authors, has not been explored before. This strategy involves the application of a GP algorithm to evolve the best performing pipelines, and conduct a statistical analysis of these pipelines.

Another contribution of our work is to show how TPOT can be extended to deal with missing data in any circumstances. Previously, only classifiers that permit \textit{missingness} on their input (e.g., XGBoost) could classify databases with missing data, but with the addition of an imputation step, any classifier can. We have shown that both,  simple and complex imputation methods, can be  added to TPOT,  giving the GP algorithm a considerably expanded search space.

As a general conclusion,  we can state that automatic configuration of pipelines can be useful to identify which are the particular combinations of imputers and classifiers more appropriate for a particular database. This result is important, since, as confirmed by previous studies, there is not a superior imputation method for all databases. 

\subsection{Further work}

This work has added new functionality to an existing python library. However, more work on the analysis part should be addressed. For example, one question to be determinedd is how the number of generations influence the quality of the optimal pipelines.


Additionally, an extension of the imputer set should be considered. In this instance, four simple methods have been proposed, complemented by two more sophisticated techniques. The latter have a significantly higher time consumption rate, which should correspond with a greater score, but this was not the case. This is why the addition of more imputation methods, whose complexity pays off, is required.

Besides, beyond the scope of this work, TPOT could be extended from a bi-objective to a tri-objective algorithm. This third component to be taken into consideration would be time. In the design we have introduced in this paper, TPOT simply focuses on obtaining the best accuracy and having short pipelines. However, pipeline shortness does not guarantee speed. Thus, pipeline execution time could be added as a new objective to optimize

Finally, and although GP looks naturally suited for this task, other optimization algorithms could be tested in this problem.

\section{Acknowledgments}

	This work was supported by IT-609-13 program (Basque Government) and TIN2016-78365-R (Spanish Ministry of Economy, Industry and Competitiveness) projects, while Unai Garciarena holds a predoctoral grant from the University of the Basque Country. We also thank Borja Calvo for useful advice on the statistical analysis conducted in the paper.
	Additionally, we thank anonymous reviewers of a previous version of this paper submitted to the SSBSE-2017 conference. Some of their comments and suggestions have been added to this manuscript.

\bibliographystyle{abbrv}

\end{document}